\documentclass[lettersize,journal]{IEEEtran}
\usepackage{amsmath,amsfonts}
\usepackage{algorithmic}
\usepackage{algorithm}
\usepackage{array}
\usepackage[caption=false,font=normalsize,labelfont=sf,textfont=sf]{subfig}
\usepackage{booktabs}
\usepackage{textcomp}
\usepackage{stfloats}
\usepackage{url}
\usepackage{verbatim}
\usepackage{graphicx}
\usepackage{cite}
\usepackage{tabularx}
\begin{document}

\title{CHESTNUT: A QoS Dataset for Mobile Edge Environments}

\author{Guobing Zou, Fei Zhao, Shengxiang Hu

\IEEEcompsocitemizethanks{
  \IEEEcompsocthanksitem G. Zou, S. Lin, S. Hu, S. Duan are with the School of Computer Engineering and Science, Shanghai University, Shanghai 200444, China.\protect\\
  E-mail: gbzou@shu.edu.cn, shengxianghu@shu.edu.cn
}
\thanks{}
}

\maketitle

\begin{abstract}
Quality of Service (QoS) is an important metric to measure the performance of network services. Nowadays, it is widely used in mobile edge environments to evaluate the quality of service when mobile devices request services from edge servers. QoS usually involves multiple dimensions, such as bandwidth, latency, jitter, and data packet loss rate. However, most existing QoS datasets, such as the common WS-Dream dataset, focus mainly on static QoS metrics of network services and ignore dynamic attributes such as time and geographic location. This means they should have detailed the mobile device's location at the time of the service request or the chronological order in which the request was made. However, these dynamic attributes are crucial for understanding and predicting the actual performance of network services, as QoS performance typically fluctuates with time and geographic location. To this end, we propose a novel dataset that accurately records temporal and geographic location information on quality of service during the collection process, aiming to provide more accurate and reliable data to support future QoS prediction in mobile edge environments.
\end{abstract}

\begin{IEEEkeywords}
Quality of service, mobile edge environments, QoS datasets, time and geographic location
\end{IEEEkeywords}

\IEEEPARstart{I}{n} the digital era, Quality of Service (QoS)\cite{kritikos2009requirements} has emerged as a fundamental concern in the realm of network services. It is influenced by various factors, including service deployment conditions, user request locations, and the adaptability of the network environment\cite{syu2017time}. As users' expectations for online services continue to rise---especially in real-time applications---QoS has a direct impact on user experience and satisfaction. Consequently, research on effective monitoring, prediction, and optimization of QoS has become a critical focus for both academia and industry. Key metrics of QoS, such as latency, bandwidth, packet loss rate, and availability, collectively provide a comprehensive evaluation of service performance\cite{karakus2017quality, electronics13163113}. In the field of web services, numerous researchers have dedicated themselves to predicting and enhancing QoS through diverse algorithms to ensure stable and reliable services, particularly under high-load conditions\cite{xue2020edge}. As the number of users and request frequencies increases, traditional web service architectures often encounter challenges such as high latency and insufficient bandwidth, which adversely affect user experience\cite{karimi2017qos}. Therefore, QoS is essential for web services, playing a vital role in improving user satisfaction and system efficiency\cite{electronics13163113}. Accurate QoS prediction can assist service providers in optimizing resource allocation\cite{Menasc2002QoSII, Shade2012QualityOS}.

Collaborative filtering has become a prominent technique for predicting missing QoS values, typically categorized into memory-based and model-based approaches. Memory-based methods leverage historical QoS invocation data from user devices, evaluating similarities between users or services to form comparable neighborhoods for predicting missing QoS values\cite{Wu2013PredictingQO, Wu2016TimeAwareAS, Li2017TemporalIC, Zheng2020WebSQ}. However, these approaches often struggle with data sparsity, making it challenging to compute accurate similarities and thus impacting prediction quality. To mitigate this issue, model-based collaborative filtering aims to develop models from historical records, extracting latent semantic features of users and services to enhance prediction accuracy. Various studies have introduced innovative model-based approaches to tackle sparsity in QoS prediction for web services. For instance, Chen Wu et al.\cite{Wu2016TimeAwareAS} proposed a time-aware and sparsity-tolerant method that combines historical QoS data with collaborative filtering, while Mingdong Tang et al.\cite{Tang2015CloudSQ} employed location-based data smoothing to enhance prediction accuracy. Additionally, Kai Su et al.\cite{Su2016WebSQ} developed a hybrid approach that integrates both memory-based and model-based techniques, utilizing neighbor information to address sparse data challenges. Qiong Zhang et al.\cite{Zhang2011CollaborativeFB} introduced a service ranking mechanism based on collaborative filtering, leveraging invocation and query histories to infer user behavior and calculate similarities. Temporal factors have also been integrated into various models to account for QoS fluctuations over time, with approaches such as Temporal Reinforced Collaborative Filtering (TRCF)\cite{Zou2024TRCFTR} and dynamic graph neural collaborative learning showcasing superior performance in time-aware QoS prediction.

Despite these advancements, collaborative filtering-based approaches still face significant challenges related to data sparsity and the inability to effectively incorporate temporal and spatial contextual information\cite{Wu2017CollaborativeQP, Zhou2023SpatialCT}. QoS performance is often influenced by multiple factors---such as time of day, user location, and network conditions---yet traditional collaborative filtering techniques fail to fully leverage this contextual information for more accurate predictions. This underscores the urgent need for exploring more advanced algorithms that can integrate these dynamic factors to enhance the effectiveness and accuracy of QoS prediction.

In recent years, there has been a notable increase in deep learning-based methods for QoS prediction in service recommendation. These approaches aim to improve prediction accuracy by effectively capturing the complex relationships among users and services. For example, Zhang et al.\cite{Zhang2023PredictingQO} introduced a two-stream deep learning model utilizing user and service graphs, while Jin et al.\cite{Jin2019NeighborhoodawareWS} developed a neighborhood-aware deep learning approach incorporating top-k neighbors. Additionally, Awanyo and Guermouche\cite{Awanyo2023DeepNN} proposed a deep neural network-based method for IoT service QoS prediction that combines Long Short-Term Memory (LSTM) networks\cite{graves2012long} for service representation with ResNet for improved accuracy\cite{he2016deep}. Zou et al.\cite{zou2022ncrl} proposed a two-tower deep neural network with residual connections to extract similar features of users and services, enhancing adaptive QoS prediction by merging deep learning with collaborative filtering.

Recent research has also shifted towards developing distributed QoS prediction models that prioritize data privacy and security while maintaining accuracy. The DISTINQT framework\cite{Koursioumpas2024DISTINQTAD} encodes raw input data into non-linear latent representations, achieving performance comparable to centralized models. The FHC-DQP framework\cite{Zou2023FHCDQPFH} integrates federated learning with hierarchical clustering to enhance prediction accuracy while preserving user privacy. Zou et al. proposed a novel model called Federated Residual Ladder Network (FRLN)\cite{zou2024frln} aimed at QoS prediction with a focus on data protection, combining the strengths of federated learning and residual networks to share learning results across multiple nodes while safeguarding user data privacy.

Given this context, it is crucial to efficiently and accurately predict QoS. Currently, researchers primarily rely on the WS-DREAM dataset\cite{zheng2012investigating} for model training and validation; however, this dataset is not well-suited for mobile edge environments. The QoS of edge devices may change dynamically as users move, contrasting with static predictions based on traditional web services, which makes it challenging for conventional methods to handle the complexities inherent in edge computing.

With the rapid advancement of 5G and cloud-edge environments, the rise of Everything-as-a-Service (XaaS)\cite{Soldani20145GNE} has facilitated a gradual transition of services from the cloud to the edge. This shift significantly enhances service flexibility and responsiveness, enabling edge devices to communicate and compute services directly with edge systems, thereby avoiding the latency associated with routing through backhaul links to remote cloud servers. As a result, network latency is drastically reduced, providing a smoother user experience but also imposing greater demands on QoS prediction.

In this dynamic landscape, QoS prediction in mobile edge environments becomes particularly critical. The variability of edge devices and the diversity of user behaviors challenge traditional QoS prediction models to adapt to new requirements. Thus, there is an urgent need for a specialized QoS prediction dataset tailored for mobile edge environments to support more accurate model training and validation. This initiative will not only improve prediction accuracy but also lay a solid foundation for future research and further promote the development and application of edge computing technologies.

Therefore, we have constructed a QoS dataset specifically for mobile edge environments, addressing the need for QoS prediction in dynamic settings. This dataset incorporates user mobility, edge server resource load, and service diversity, providing a more realistic simulation of the operational environment. Additionally, we define a paradigm for calculating QoS values based on the unique characteristics of edge environments, thus establishing a robust data foundation for subsequent QoS prediction models in mobile edge scenarios. By utilizing this dataset, researchers can more effectively develop and validate models to tackle the intricate challenges posed by mobile edge computing. To facilitate reproducible research, the public release of CHESTNUT is available on Kaggle at \url{https://www.kaggle.com/datasets/sorriso07/chestnut}.

\section{original dataset description}
\label{sec:description}

To create a high-quality dataset for predicting Quality of Service (QoS) in mobile edge environments, this study utilized two real-world datasets from Shanghai. One dataset is from the Shanghai Johnson Taxi, containing information such as the longitude, latitude, moving direction, and speed of the taxis on a specific day, which was used to simulate user mobile datasets. The other dataset is from Shanghai Telecom \cite{li2021profit, guo2020user, wang2019delay}, providing the longitude and latitude of the base stations. Data from June 2014 was used to simulate the generation of edge server datasets.

\subsection{Shanghai Johnson Taxi Dataset}
The Shanghai johnson taxi dataset is an important resource for traffic research, containing real-time GPS and business status information from taxis in Shanghai. Each record in the dataset includes various fields: the vehicle ID, control word (where A indicates normal and M indicates alarm), business status (0 for normal and 1 for alarm), passenger status (0 for occupied and 1 for unoccupied), top light status (with values ranging from 0 for operation to 5 for out of service), road type (0 for ground road and 1 for express road), brake status (0 for no braking and 1 for braking), meaningless fields, reception date, GPS timestamp, longitude, latitude, speed, direction, the number of satellites, and additional meaningless fields. This dataset enables the analysis of urban traffic flow, travel patterns, and traffic management strategies. In this paper, we use only the gps time, latitude, longitude, speed, and direction of the cab to generate motion information about the mobile user.

\subsection{Shanghai Telecom Dataset}
This study utilized a telecommunications dataset provided by Shanghai Telecom, comprising over 7.2 million records of 9,481 mobile phones accessing the Internet through 3,233 base stations over six months.\footnote{\url{http://sguangwang.com/TelecomDataset.html}} The dataset includes six parameters: month, date, start time, end time, base station location (latitude and longitude), and user ID used within Shanghai Telecom.

This dataset can be used to evaluate solutions in mobile edge computing, such as edge server deployment, service migration, and service recommendation. In our research, we need to simulate the information of edge servers on base stations based on this dataset. Furthermore, considering the substantial volume of data, we only counted the data for one month and only focused on the geographic location information of Shanghai base stations.

\section{Data Preparation}

CHESTNUT is built by integrating the two real-world Shanghai datasets above and then synthesizing the missing edge-side attributes required for mobile QoS research. Taxi trajectories are used to emulate mobile users, while telecom base stations are treated as candidate edge servers with fixed locations. Because the raw datasets were collected for traffic analysis and network operation rather than QoS modeling, several preprocessing steps are required before they can be used to generate user traces, server states, service types, and invocation records. The main configuration is summarized in Table \ref{tab:table1}.
\begin{table}[ht]
    \centering
    \caption{Simulation Parameters}
    \label{tab:table1}
    \tabcolsep=0.42cm
    \begin{tabular}{ll}
        \toprule
        \textbf{Parameter} & \textbf{Value} \\
        \midrule
        Minimal latitude $\phi_-$ & 31.050 \\
        Maximal latitude $\phi_+$ & 31.372 \\
        Minimal longitude $\lambda_-$ & 121.259 \\
        Maximal longitude $\lambda_+$ & 121.640 \\
        Minimal server coverage radius $r_-$ & 400 \\
        Maximal server coverage radius $r_+$ & 3200 \\
        Number of users $N_u$ & 1000 \\
        Number of services $N_s$ & 1000 \\
        Maximal server resource level $P_e$ & 6 \\
        Maximal service demand level $P_s$ & 4 \\
        Time alignment interval $\Delta t$ & 30 \\
        Maximum system time $T_{max}$ & 21600 \\
        Minimum valid timestamps $C_{min}$ & 30 \\
        Maximum stationary interval $S$ & 3 \\
        Base delay $\Theta_{RT}$ & 1.6 \\
        Base jitter $\Theta_{NJ}$ & 160 \\
        History window size $k$ & 5 \\
        \bottomrule
    \end{tabular}
\end{table}

\subsection{Edge Server Data}
\begin{figure}[htbp]
\centering
\includegraphics[scale=0.26]{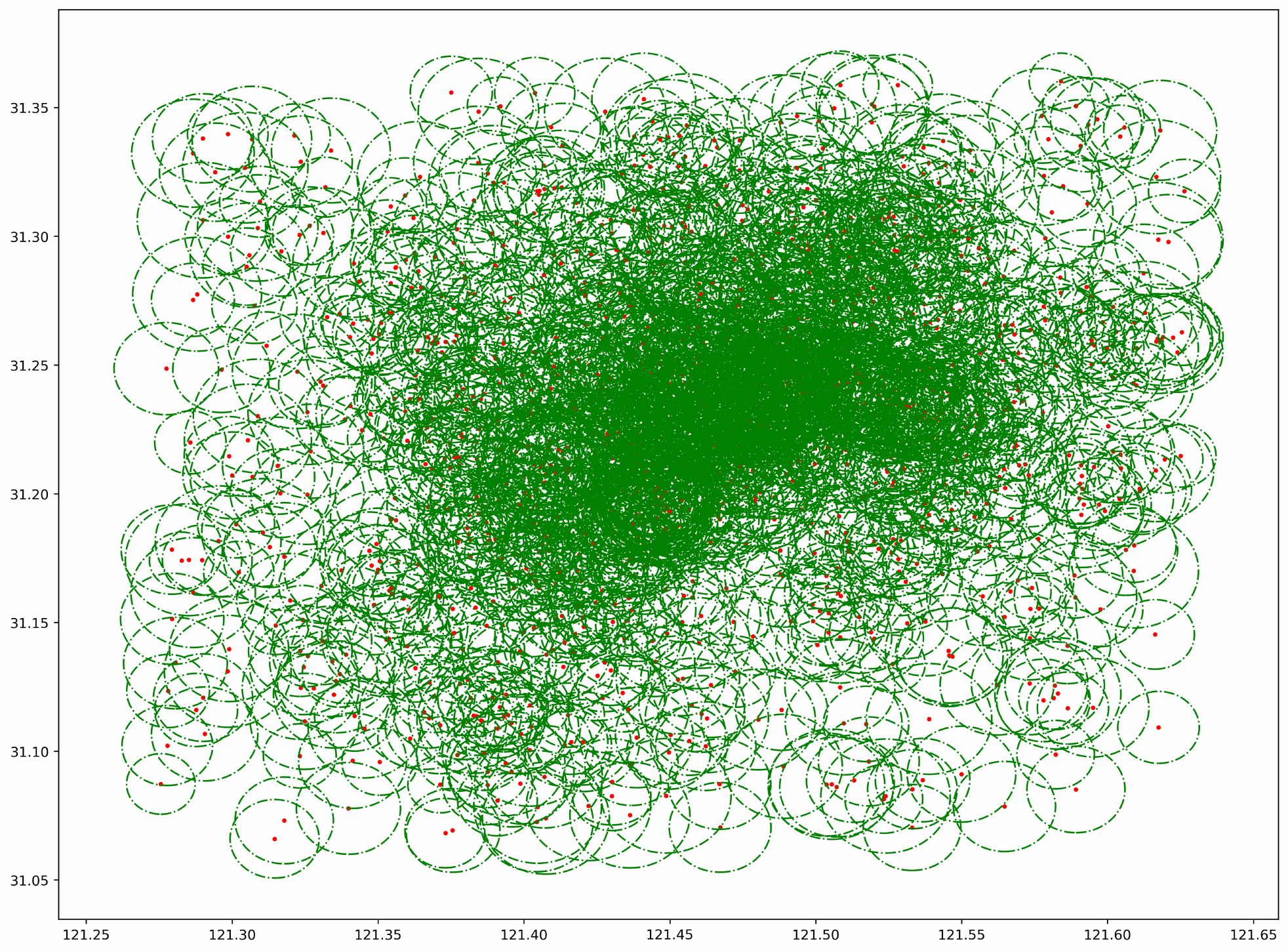}
\caption{Edge server distribution in the selected Shanghai area.}
\label{fig:figure1}
\end{figure}

Edge server locations are extracted from the Shanghai Telecom dataset \cite{li2021profit, guo2020user, wang2019delay}. We keep only the base stations whose coordinates fall inside the target urban area bounded by $[\phi_-, \phi_+]$ and $[\lambda_-, \lambda_+]$. This produces a geographically dense deployment, which is important because mobile users should be able to move across overlapping coverage regions rather than frequently falling into uncovered zones. The resulting server map is illustrated in Figure \ref{fig:figure1}.

The telecom records provide base station coordinates but not radio coverage radii. To approximate heterogeneous 4G and 5G edge access ranges, each selected site is assigned a radius sampled from $[r_-, r_+]$. Since geographic coverage is measured on the Earth surface rather than on a plane, we compute distances with the Haversine formula:
\begin{align}
\operatorname{hav}(\theta) &= \operatorname{hav}(\Delta \phi) + \cos(\phi_1)\cos(\phi_2)\operatorname{hav}(\Delta \lambda) \\
D &= 2R \arcsin \left(\sqrt{\operatorname{hav}(\theta)}\right)
\end{align}
where $R$ is the Earth radius, $\Delta \phi = \phi_2 - \phi_1$, and $\Delta \lambda = \lambda_2 - \lambda_1$.

Each server is further assigned three resource supply levels for computation, storage, and bandwidth. These values are integers in $[1, P_e]$, where larger values indicate stronger provisioning capacity. Instead of fixing device-specific hardware specifications, CHESTNUT models relative resource strength, which is sufficient for downstream QoS prediction and makes the dataset easier to generalize. A server $e_i$ is therefore represented as
\begin{align}
e_i = \langle i, \lambda_i, \phi_i, r_i, \psi_{i,c}, \psi_{i,s}, \psi_{i,b} \rangle
\end{align}

\subsection{Service Data}
To model heterogeneous edge applications without binding the dataset to a particular software stack, CHESTNUT synthesizes $N_s$ service types. Each service is described by three demand levels: computation, storage, and bandwidth. The three values are sampled from $[1, P_s]$, where a larger value means the service consumes more of the corresponding resource. Although at most $P_s^3$ unique demand combinations exist, services with the same demand tuple are still retained as different service IDs to mimic latent application differences that are not explicitly represented in the dataset.

This design keeps the service schema compact while preserving enough heterogeneity for QoS analysis. A sample of the service table is shown in Table \ref{tab:table3}.
\begin{table}[ht]
	\caption{Sample Service Records}
	\label{tab:table3}
	\centering
	\tabcolsep=0.5cm
	\begin{tabular}{cccc}
	\toprule
	\textbf{sid} & \textbf{computing} & \textbf{storage} & \textbf{bandwidth} \\
	\midrule
	0 & 1 & 4 & 2 \\
	1 & 4 & 2 & 3 \\
	2 & 3 & 1 & 4 \\
	3 & 2 & 3 & 1 \\
	4 & 4 & 4 & 2 \\
	\bottomrule
	\end{tabular}
\end{table}

\subsection{User Data}
The Shanghai Johnson Taxi dataset is used to emulate mobile users. After removing irrelevant columns, each retained record contains the taxi ID, GPS timestamp, longitude, latitude, instantaneous speed, and moving direction. These fields are enough to reconstruct fine-grained user mobility, but they still need to be aligned and filtered because taxis become active at different times and report GPS records at irregular intervals.

\subsubsection{Temporal Alignment}
\begin{figure}[htbp]
\centering
\includegraphics[scale=0.28]{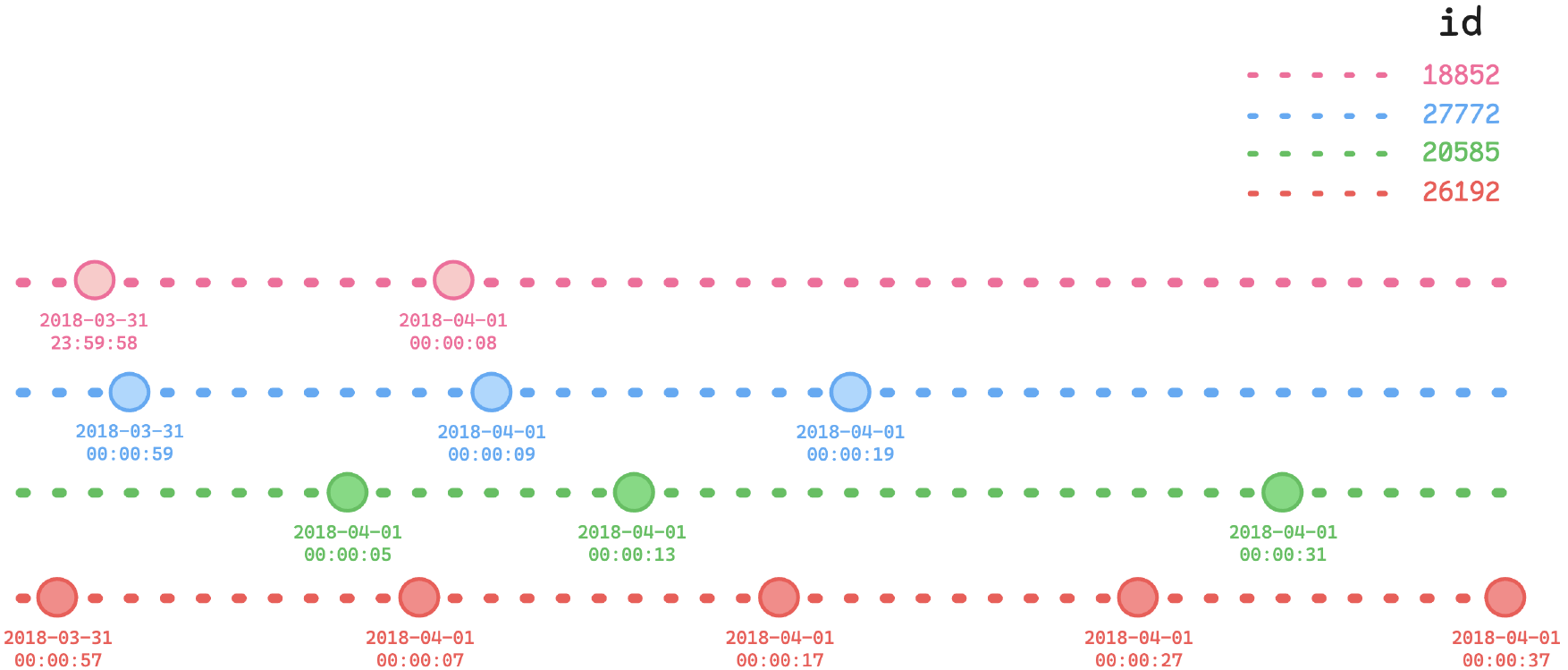}
\caption{Examples of irregular taxi GPS sequences before alignment.}
\label{fig:figure2}
\end{figure}

The first issue is that the original trajectories are not synchronized. In CHESTNUT, the first GPS record of each taxi is mapped to the initial system timestamp, and the trajectory is then projected onto a discrete edge-system clock. Figure \ref{fig:figure2} shows the irregular reporting pattern in the raw data.

\begin{figure}[htbp]
\centering
\includegraphics[scale=0.28]{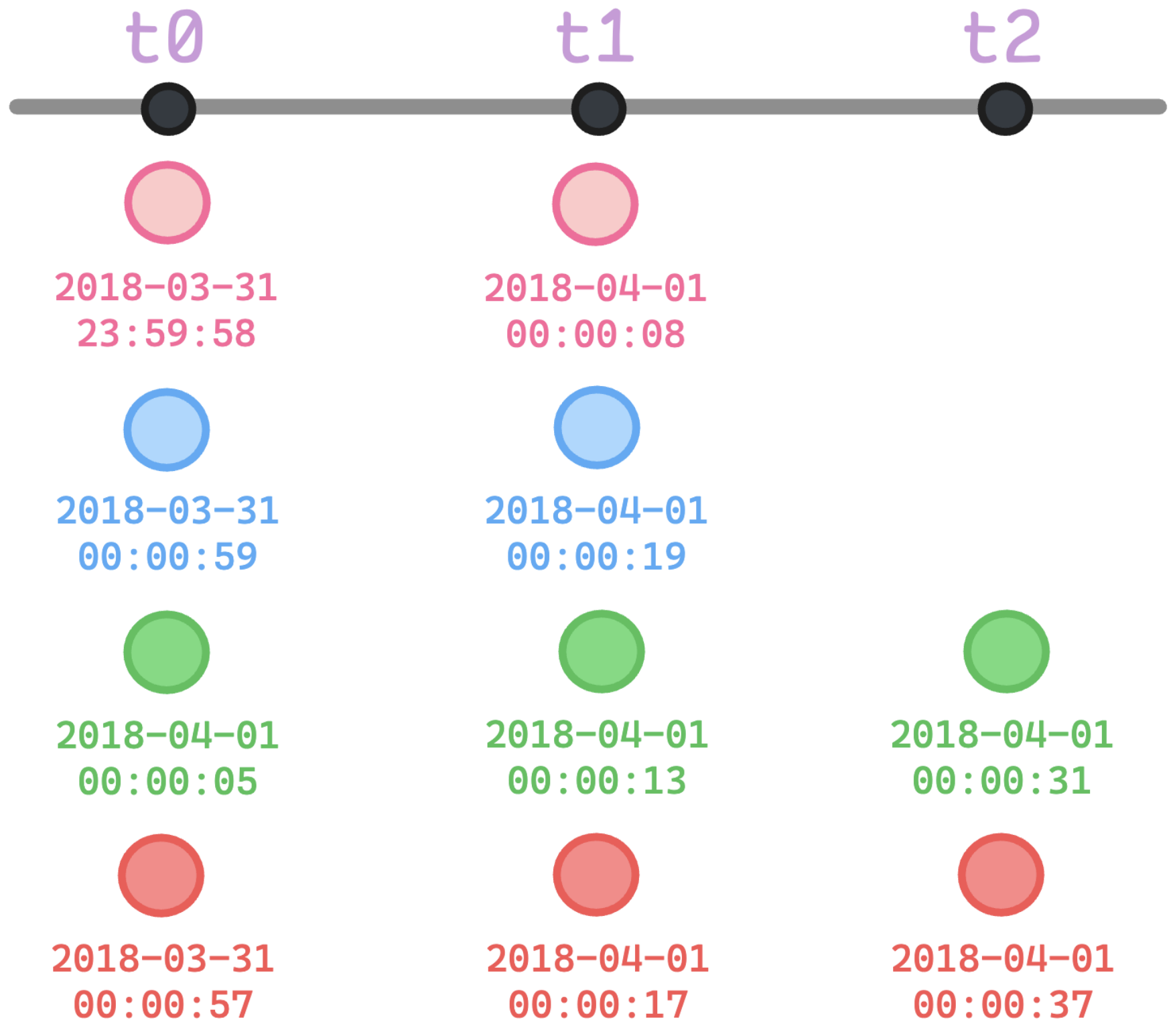}
\caption{Aligned user trajectories after fixed-window snapshotting.}
\label{fig:figure4}
\end{figure}

To reduce temporal inconsistency, we further partition each trajectory into fixed windows of length $\Delta t$ seconds. If several records fall into the same window, only the last one is kept as the valid snapshot for that timestamp. If no record appears in a window, the user is treated as temporarily absent from the monitored edge environment. The aligned result is illustrated in Figure \ref{fig:figure4}.

For a user $u$, suppose the aligned trajectory contains $\tau_u$ valid timestamps. We denote its timestamp set by $K_u = \{\kappa_1, \kappa_2, \ldots, \kappa_{\tau_u}\}$, where
\begin{align}
\kappa_i \in \left[0, \left\lfloor \frac{T_{max}}{\Delta t} \right\rfloor \right].
\end{align}
This procedure yields user traces that are temporally comparable across the whole system.

\subsubsection{Active User Selection}
Even after alignment, many taxis contribute little useful motion diversity because they stay nearly static for long periods or appear only briefly. Such traces are weak supports for mobile QoS prediction, so we keep only the most informative users. First, we remove users whose longest stationary streak exceeds $S$ timestamps or whose total number of valid timestamps is below $C_{min}$. Then, for each remaining user $u$, we compute four mobility-related indicators:
\begin{itemize}
    \item $\tau_u$: the number of valid timestamps.
    \item $D_u$: the great-circle distance between the first and last observed locations.
    \item $\omega_u^t$: the number of servers covering user $u$ at timestamp $t$.
    \item $\nu_u^t$: the number of timestamps at which user $u$ is covered by at least one server.
\end{itemize}

\begin{figure}[htbp]
\centering
\includegraphics[scale=0.26]{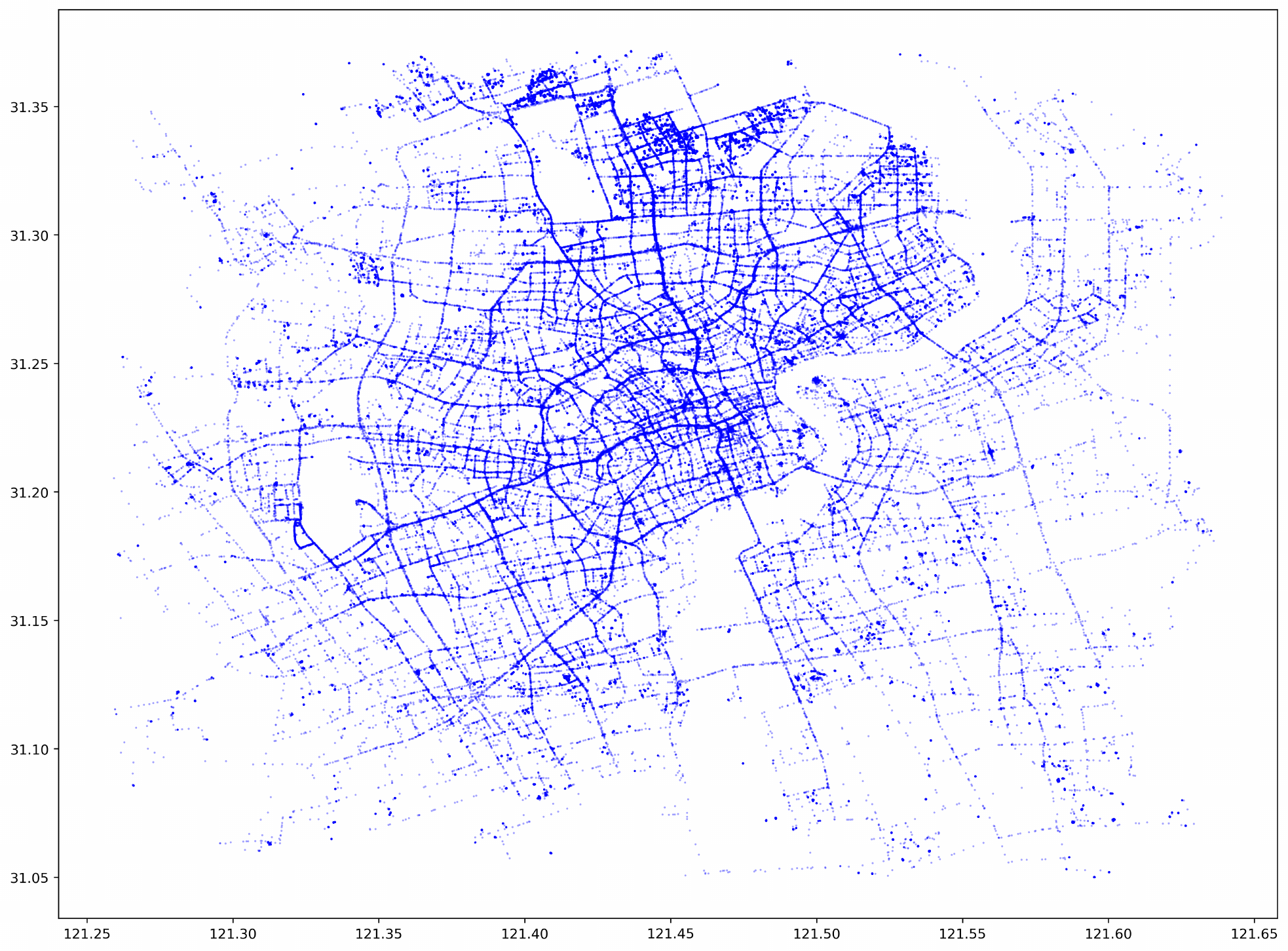}
\caption{Spatial distribution of the selected mobile users.}
\label{fig:figure5}
\end{figure}

\begin{figure}[htbp]
\centering
\includegraphics[width=0.95\columnwidth]{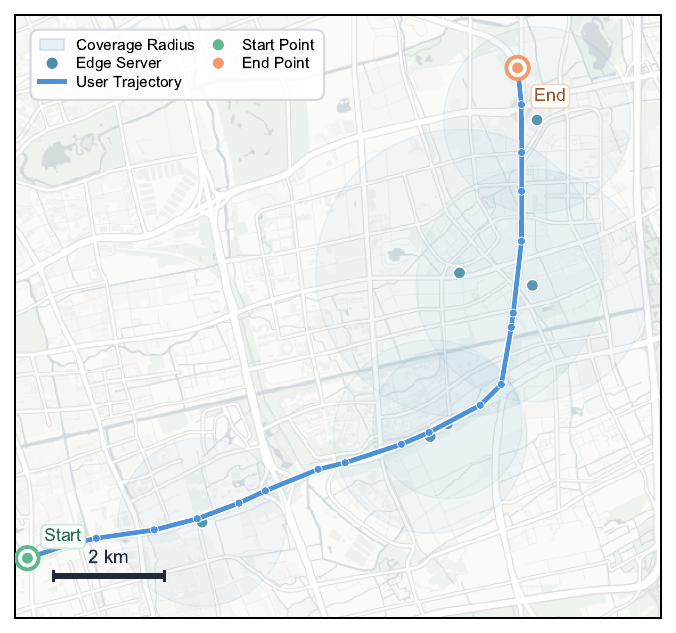}
\caption{A local example of user movement and overlapping edge coverage.}
\label{fig:trajectory_coverage_map}
\end{figure}

Users are ranked in descending lexicographic order by
\begin{align}
\left(\tau_u, D_u, \sum_{t \in T_u} \omega_u^t, \sum_{t \in T_u} \nu_u^t\right),
\end{align}
which favors long-lived, highly mobile users that frequently move through densely covered areas. We then keep the top $N_u$ users and reindex them. Figure \ref{fig:figure5} shows the resulting user distribution, while Figure \ref{fig:trajectory_coverage_map} gives a local example of how a moving user traverses multiple overlapping server regions.

\begin{table*}[ht]
	\caption{Sample User Records}
	\label{tab:table4}
	\centering
	\tabcolsep=0.6cm
	\begin{tabular}{cccccc}
	\toprule
	\textbf{id} & \textbf{timestamp} & \textbf{lon} & \textbf{lat} & \textbf{speed} & \textbf{direction} \\
	\midrule
	0 & 0 & 121.431827 & 31.306520 & 0.0 & 84 \\
	1 & 0 & 121.478397 & 31.321983 & 0.0 & 305 \\
	1 & 1 & 121.478383 & 31.321997 & 0.0 & 305 \\
	91 & 3 & 121.469835 & 31.285843 & 0.0 & 0 \\
	77 & 7 & 121.334412 & 31.264438 & 20.0 & 180 \\
	\bottomrule
	\end{tabular}
\end{table*}

\section{Synthetic Data Generation}
After obtaining aligned user traces, server attributes, and service types, we synthesize dynamic service invocations and their QoS values. The generation process models not only where a user is located, but also how often it requests services, which server receives the requests, how server loads evolve, and how latency and jitter emerge from user-server interactions. Table \ref{tab:table5} lists the main symbols used below.

\begin{table}[ht]
    \centering
    \renewcommand{\arraystretch}{1.2} 
    \caption{Main Notations}
    \label{tab:table5}
    \begin{tabularx}{\linewidth}{lX} 
        \toprule
        \textbf{Notation} & \textbf{Description} \\
        \midrule
        $\varphi_{s,c}, \varphi_{s,s}, \varphi_{s,b}$ & service demand levels for computation, storage, and bandwidth \\
        $\psi_{e,c}, \psi_{e,s}, \psi_{e,b}$ & server supply levels for computation, storage, and bandwidth \\
        $R_e$ & coverage radius of server $e$ \\
        $\rho_{e,c}^t, \rho_{e,s}^t, \rho_{e,b}^t$ & resource load rates of server $e$ at timestamp $t$ \\
        $q_u^t$ & number of requests issued by user $u$ at timestamp $t$ \\
        $v_u^t$ & instantaneous speed of user $u$ at timestamp $t$ \\
        $\theta_u^t$ & moving direction of user $u$ at timestamp $t$ \\
        \bottomrule
    \end{tabularx}
\end{table}

\subsection{Invocation Sequence Synthesis}
In real mobile edge systems, different users often exhibit correlated but non-identical request rhythms. To reflect this, we divide users into several request-pattern groups and assign each group a shared activity timeline. For a group $g_i$, all users in that group inherit the same active timestamp set $T_{g_i}$, which captures coarse-grained common behaviors such as commuting peaks or periodic bursts.

Within the shared timeline, each user receives an individualized request-count signal generated by a Gaussian process \cite{rasmussen2006gaussian} with additive white noise. For user $u$,
\begin{align}
f_u(t) &\sim \mathcal{GP}(0, k(t,t')) \\
\epsilon_u(t) &\sim \mathcal{N}(0,\sigma_u^2), \qquad \sigma_u \sim \mathcal{U}(0.1,0.3) \\
s_u(t) &= f_u(t) + \epsilon_u(t)
\end{align}
where the kernel $k(t,t')$ is selected from a generator pool containing smooth, periodic, and bursty temporal patterns. The continuous signal is then shifted and scaled into a non-negative integer request count:
\begin{align}
q_u^t = \left\lceil 20 \cdot \left(s_u(t) - \min_{\tau \in T_u} s_u(\tau)\right)\right\rceil
\end{align}

Each of the $q_u^t$ requests is sent to the nearest server that covers user $u$ at timestamp $t$. The target service is sampled from the service subset associated with the user's request-pattern group. In this way, users in the same group share similar global trends while still exhibiting user-specific local variations. An invocation record is defined as
\begin{align}
\mathcal{R}_i = \langle u, e, s, t \rangle
\end{align}
and all records together form the invocation set $\mathcal{R}$.

\subsection{Edge Server Load}
Server loads should evolve continuously rather than being regenerated independently at each timestamp. For this reason, CHESTNUT uses a recursive load model. At the beginning of the system, each server is assigned an initial background load vector $\rho_e^0 = [\rho_{e,c}^0, \rho_{e,s}^0, \rho_{e,b}^0]$. At timestamp $t-1$, the remaining supply levels of server $e$ are
\begin{align}
p_{e,c}^{t-1} &= (1-\rho_{e,c}^{t-1}) \cdot \psi_{e,c} \\
p_{e,s}^{t-1} &= (1-\rho_{e,s}^{t-1}) \cdot \psi_{e,s} \\
p_{e,b}^{t-1} &= (1-\rho_{e,b}^{t-1}) \cdot \psi_{e,b}
\end{align}

Let $\mathcal{B}_{t-1}^e$ be the set of services assigned to server $e$ at timestamp $t-1$. We normalize the current supply vector and the aggregated demand vector separately:
\begin{align}
\alpha_e^{t-1} &= \operatorname{Softmax}\left([p_{e,c}^{t-1}, p_{e,s}^{t-1}, p_{e,b}^{t-1}]\right) \\
\beta_e^{t-1} &= \operatorname{Softmax}\left(\left[\sum_{s \in \mathcal{B}_{t-1}^e}\varphi_{s,c}, \sum_{s \in \mathcal{B}_{t-1}^e}\varphi_{s,s}, \sum_{s \in \mathcal{B}_{t-1}^e}\varphi_{s,b}\right]\right)
\end{align}
The discrepancy between demand and supply is then injected into the previous load state:
\begin{align}
\gamma_e^{t-1} = \operatorname{Softmax}\left(\operatorname{Softmax}(\beta_e^{t-1}-\alpha_e^{t-1}) + [\rho_{e,c}^{t-1}, \rho_{e,s}^{t-1}, \rho_{e,b}^{t-1}]\right)
\end{align}

To account for task completion and resource release, we multiply the intermediate load by a nonlinear release factor $\eta = \tanh(x)$ with $x \sim \mathcal{U}(0,1)$:
\begin{align}
\rho_e^t = \gamma_e^{t-1} \cdot \eta
\end{align}
This produces temporally smooth load series with both accumulation and decay. Sample load records are shown in Table \ref{tab:table6}.

\begin{table*}[ht]
	\caption{Sample Server Load Records}
	\label{tab:table6}
	\centering
	\tabcolsep=0.6cm
	\begin{tabular}{ccccc}
	\toprule
	\textbf{timestamp} & \textbf{eid} & \textbf{computing\_load} & \textbf{storage\_load} & \textbf{bandwidth\_load} \\
	\midrule
	0 & 0 & 14.78\% & 6.45\% & 5.67\% \\
	140 & 625 & 97.63\% & 11.00\% & 72.26\% \\
	161 & 1089 & 6.33\% & 16.24\% & 11.15\% \\
	213 & 1270 & 54.09\% & 16.01\% & 50.13\% \\
	320 & 1031 & 34.21\% & 7.72\% & 93.04\% \\
	\bottomrule
	\end{tabular}
\end{table*}

\subsection{Response Time Data Generation}
The response time of an invocation is modeled as the sum of six stages: request propagation delay, uplink transmission delay, queueing delay, processing delay, downlink transmission delay, and response propagation delay \cite{Chen2008AnalysisOW, Ruan2017RoundTripDM, Ulvan2009DelayPO, kurose2017computer}. This decomposition makes it possible to link latency to both network geometry and server-side resource dynamics.

\subsubsection{Request Propagation Delay}
The request propagation delay depends on the great-circle distance between the user and the selected server:
\begin{align}
PG_i^{req} = \frac{\operatorname{Haversine}(\lambda_u^t,\phi_u^t,\lambda_e,\phi_e)}{c}
\end{align}
where $c = 3 \times 10^8$ m/s is the propagation speed of radio waves.

\subsubsection{Transmission Delay}
Bandwidth supply levels are mapped to a discrete bandwidth set $\mathcal{W}_e = \{16, 64, 128, 512, 1024, 2048\}$ Mbps, while service bandwidth demand levels are mapped to packet sizes $\mathcal{W}_s = \{1, 8, 64, 512, 2048, 4096\}$ KB. For a service $s$ assigned to server $e$ at timestamp $t$, we define
\begin{align}
B_{e,s}^t &= W_{\lfloor \psi_{e,b} \rfloor}^e \cdot (1-\rho_{e,b}^t) \cdot \xi \cdot \frac{\varphi_{s,b}}{\sum_{s' \in \mathcal{B}_t^e}\varphi_{s',b}} \\
L_s &= (1 + \varphi_{s,b} - \lfloor \varphi_{s,b} \rfloor) \cdot W_{\lfloor \varphi_{s,b} \rfloor}^s
\end{align}
where $\xi = 128$ is the conversion factor from Mbps to KB/s. The uplink and downlink transmission delays are
\begin{align}
U_i &= \frac{L_s}{B_{e,s}^t} \\
D_i &= \frac{L_s}{W_{\lfloor \psi_{e,b} \rfloor}^e \cdot (1-\rho_{e,b}^t) \cdot \xi}
\end{align}

\subsubsection{Queueing Delay}
Before execution, each invocation waits for computation, storage, and bandwidth resources. We model the three waiting stages with M/M/1 queues \cite{kleinrock1975queueing}. To better separate low-demand and high-demand services, the resource demand levels are first expanded by
\begin{align}
\tilde{\varphi}_{s,*} = 1.5^{\varphi_{s,*}}, \qquad * \in \{c,s,b\}.
\end{align}
For each resource type,
\begin{align}
\lambda_{e,s,*}^t &= \frac{\tilde{\varphi}_{s,*}}{\sum_{s' \in \mathcal{B}_t^e}\tilde{\varphi}_{s',*}} \\
\mu_{e,s,*}^t &= \lambda_{e,s,*}^t \cdot \left(\psi_{e,*}(1-\rho_{e,*}^t)+1\right) \\
\chi_{e,s,*}^t &= \frac{\psi_{e,*}(1-\rho_{e,*}^t)}{\tilde{\varphi}_{s,*}} \\
Q_{i,*} &= e^{-\chi_{e,s,*}^t} \cdot \frac{\lambda_{e,s,*}^t}{\mu_{e,s,*}^t(\mu_{e,s,*}^t - \lambda_{e,s,*}^t)}
\end{align}
The total queueing delay is $Q_i = Q_{i,c} + Q_{i,s} + Q_{i,b}$.

\subsubsection{Processing Delay}
After the required resources become available, service execution is governed mainly by computation and storage:
\begin{align}
P_i = \frac{\tilde{\varphi}_{s,c}}{(1-\rho_{e,c}^t)\psi_{e,c}} \cdot (1+\rho_{e,c}^t) + \frac{\tilde{\varphi}_{s,s}}{(1-\rho_{e,s}^t)\psi_{e,s}} \cdot (1+\rho_{e,s}^t)
\end{align}
This form penalizes both high demand and high server utilization.

\subsubsection{Delay Regularization}
The transmission, queueing, and processing terms are generated by different abstractions and therefore lie on different scales. To combine them without allowing extreme values to dominate, CHESTNUT first compresses them with logarithmic transforms and then applies min-max normalization followed by a $\tanh$ nonlinearity:
\begin{align}
I_i &= \log(1+U_i) + \sum_{* \in \{c,s,b\}}\log(1+Q_{i,*}) \nonumber \\
&\quad + \log(1+P_i+D_i) \\
\end{align}
Here $SD_i$ is the server-side delay component expressed in milliseconds.

\subsubsection{Response Propagation Delay}
Because the user may have moved while the request is being processed, the return packet is assumed to travel to an estimated future location rather than the original one. Let $v_u^t$ be the user speed in m/s and $\theta_u^t$ the travel direction. The estimated travel distance before the response is delivered is
\begin{align}
d = v_u^t \cdot (PG_i^{req} + SD_i).
\end{align}
The predicted receiving position is
\begin{align}
\hat{\phi}_u^t &= \phi_u^t + \frac{d \cdot \cos(\operatorname{rad}(\theta_u^t))}{111320} \\
\hat{\lambda}_u^t &= \lambda_u^t + \frac{d \cdot \sin(\operatorname{rad}(\theta_u^t))}{111320 \cdot \cos(\operatorname{rad}(\phi_u^t))}
\end{align}
and the response propagation delay is
\begin{align}
PG_i^{rep} = \frac{\operatorname{Haversine}(\lambda_e,\phi_e,\hat{\lambda}_u^t,\hat{\phi}_u^t)}{c}
\end{align}

Finally, the total response time is
\begin{align}
RT_i = \kappa \cdot PG_i^{req} + SD_i + \kappa \cdot PG_i^{rep}
\end{align}
where $\kappa = 10^4$ is a scale factor used to bring propagation delays to the same observable range as the server-side delay.

\subsection{Network Jitter Data Generation}
Network jitter captures delay fluctuation rather than absolute delay \cite{rfc3393}. In mobile edge systems, it is influenced by user mobility, bandwidth pressure, and the spatial relation between users and servers \cite{chen2023mobility, jin2024swift}. CHESTNUT combines five factors to synthesize jitter.

\subsubsection{Bandwidth Load Trend}
The first factor is the recent trend of server bandwidth utilization:
\begin{align}
\varsigma_i^t = \frac{1}{|T_e^t|}\sum_{\tau \in T_e^t}\left(\rho_{e,b}^{\tau} - \rho_{e,b}^{\tau-1}\right)
\end{align}
where $T_e^t$ contains up to $k$ historical timestamps before $t$. A positive trend indicates rising bandwidth pressure and hence a less stable transmission environment.

\subsubsection{User-Server Distance Ratio}
The second factor measures how close the user is to the boundary of the serving region:
\begin{align}
\varsigma_i^d = \frac{\operatorname{Haversine}(\lambda_u^t,\phi_u^t,\lambda_e,\phi_e)}{R_e}.
\end{align}
Larger values imply a weaker and potentially less stable connection.

\subsubsection{Average Directional Change}
The third factor quantifies short-term changes in user movement direction:
\begin{align}
\varsigma_i^c = \frac{1}{|T_u^t|}\sum_{\tau \in T_u^t} |\theta_u^{\tau} - \theta_u^{\tau-1}|
\end{align}
where $T_u^t$ contains up to $k$ historical records of user $u$.

\subsubsection{Bandwidth Demand Ratio}
The fourth factor characterizes the mismatch between service demand and currently available bandwidth:
\begin{align}
\varsigma_i^r = \frac{\varphi_{s,b}}{\psi_{e,b}(1-\rho_{e,b}^t)}
\end{align}
Large values indicate that the request consumes a large fraction of the server's remaining bandwidth.

\subsubsection{Jitter Fusion}
The fifth factor is the user's instantaneous speed $v_u^t$. After min-max normalization, the five terms are fused as
\begin{align}
\varsigma_i &= e^{1+\hat{\varsigma}_i^t} \cdot \left(\hat{\varsigma}_i^d + \hat{\varsigma}_i^c + \hat{\varsigma}_i^r + \hat{v}_i\right) \\
J_i &= \left(\tanh\left(M_{[-2,\ 2]}(\varsigma_i)\right)+1\right)\Theta_{NJ}
\end{align}
where $J_i$ is the final jitter value in milliseconds.

\subsection{Perturbations}
To avoid making QoS values overly deterministic, CHESTNUT injects two additional perturbations. The first is an edge-context perturbation generated from the triplet $(u_i,e_i,s_i)$ by a lightweight feedforward model:
\begin{align}
\delta_{edge}^{u_i,e_i,s_i} = \operatorname{Model}([u_i,e_i,s_i])
\end{align}
The second is a time perturbation:
\begin{align}
\delta_{time}^{t_i} = 0.1\left(\sin\left(\frac{t_i}{2}\right)+1\right)
\end{align}
Both perturbations are scaled into $[0, 0.2]$, so the final QoS values fluctuate within a moderate but meaningful range:
\begin{align}
Q_{rt}^i &= RT_i \cdot \left(1 + \delta_{edge}^{u_i,e_i,s_i} + \delta_{time}^{t_i}\right) \\
Q_{nj}^i &= J_i \cdot \left(1 + \delta_{edge}^{u_i,e_i,s_i} + \delta_{time}^{t_i}\right)
\end{align}

Table \ref{tab:table7} lists several sample invocation records.
\begin{table*}[ht]
    \caption{Sample Invocation Records}
    \label{tab:table7}
    \centering
    \tabcolsep=0.4cm
    \begin{tabular}{cccccc}
    \toprule
    \textbf{uid} & \textbf{eid} & \textbf{sid} & \textbf{timestamp} & \textbf{rt} & \textbf{nj} \\
    \midrule
    0 & 606 & 27 & 0 & 0.162801 & 8.963441 \\
    54 & 732 & 29 & 0 & 0.173344 & 12.486833 \\
    8 & 8 & 108 & 3 & 0.097180 & 8.400173 \\
    54 & 210 & 16 & 77 & 0.357487 & 9.464810 \\
    77 & 1689 & 79 & 111 & 0.175501 & 17.383079 \\
    \bottomrule
    \end{tabular}
\end{table*}

\section{Results and Analysis}
This section reports the main statistics of CHESTNUT and analyzes whether the generated dataset reflects the expected properties of a mobile edge environment.

\begin{table}[ht]
    \centering
    \caption{Overall Statistics of CHESTNUT}
    \label{tab:table8}
    \tabcolsep=0.75cm
    \begin{tabular}{ll}
        \toprule
        \textbf{Item} & \textbf{Value} \\
        \midrule
        Users & 1,000 \\
        Servers & 1,763 \\
        Services & 1,000 \\
        Timestamps & 720 \\
        Invocations & 33,551,574 \\
        \bottomrule
    \end{tabular}
\end{table}

As summarized in Table \ref{tab:table8}, CHESTNUT contains full four-dimensional interactions among users, servers, services, and timestamps. Compared with datasets designed for traditional web-service QoS prediction, it also exposes the spatial and dynamic server-side context required by mobile edge research. Table \ref{tab:dataset_compare} highlights the difference from WS-DREAM.

\begin{table*}[ht]
  \centering
  \caption{Comparison Between WS-DREAM and CHESTNUT}
  \label{tab:dataset_compare}
  \renewcommand{\arraystretch}{1.15}
  \tabcolsep=0.28cm
  \footnotesize
  \begin{tabular}{p{8.6cm}cc}
    \toprule
    \textbf{Property} & \textbf{WS-DREAM} & \textbf{CHESTNUT} \\
    \midrule
    User ID and service ID & $\checkmark$ & $\checkmark$ \\
    Server ID &  & $\checkmark$ \\
    Timestamp index & $\checkmark$ & $\checkmark$ \\
    User-service QoS records & $\checkmark$ & $\checkmark$ \\
    User-server-service QoS records &  & $\checkmark$ \\
    User trajectory sequence &  & $\checkmark$ \\
    Server location metadata & $\checkmark$ & $\checkmark$ \\
    User-server coverage relation &  & $\checkmark$ \\
    Server resource supply &  & $\checkmark$ \\
    Dynamic server load &  & $\checkmark$ \\
    Service demand attributes &  & $\checkmark$ \\
    Response time & $\checkmark$ & $\checkmark$ \\
    Throughput & $\checkmark$ &  \\
    Network jitter &  & $\checkmark$ \\
  \bottomrule
  \end{tabular}
\end{table*}

\subsection{User Mobility and Coverage Characteristics}
\begin{figure}[H]
\centering
\includegraphics[width=\columnwidth]{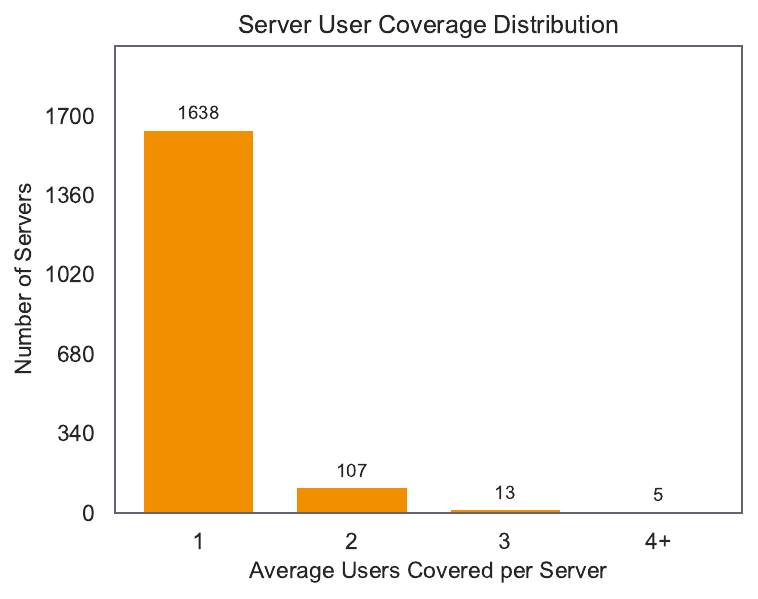}
\caption{Distribution of the average number of users covered by each server.}
\label{fig:figure6}
\end{figure}

\textbf{User continuity and server coverage.} After screening, all selected users exhibit fully continuous active timestamps under the aligned system clock, which means the final user trajectories are suitable for temporal QoS modeling rather than sparse event matching. Figure \ref{fig:figure6} reports the distribution of the average number of users covered by a server.
Most servers cover between one and three users on average, and only five servers exceed four covered users. This light but nontrivial overlap matches the spatially scattered behavior expected in urban edge scenarios.

\begin{figure}[H]
\centering
\includegraphics[width=\columnwidth]{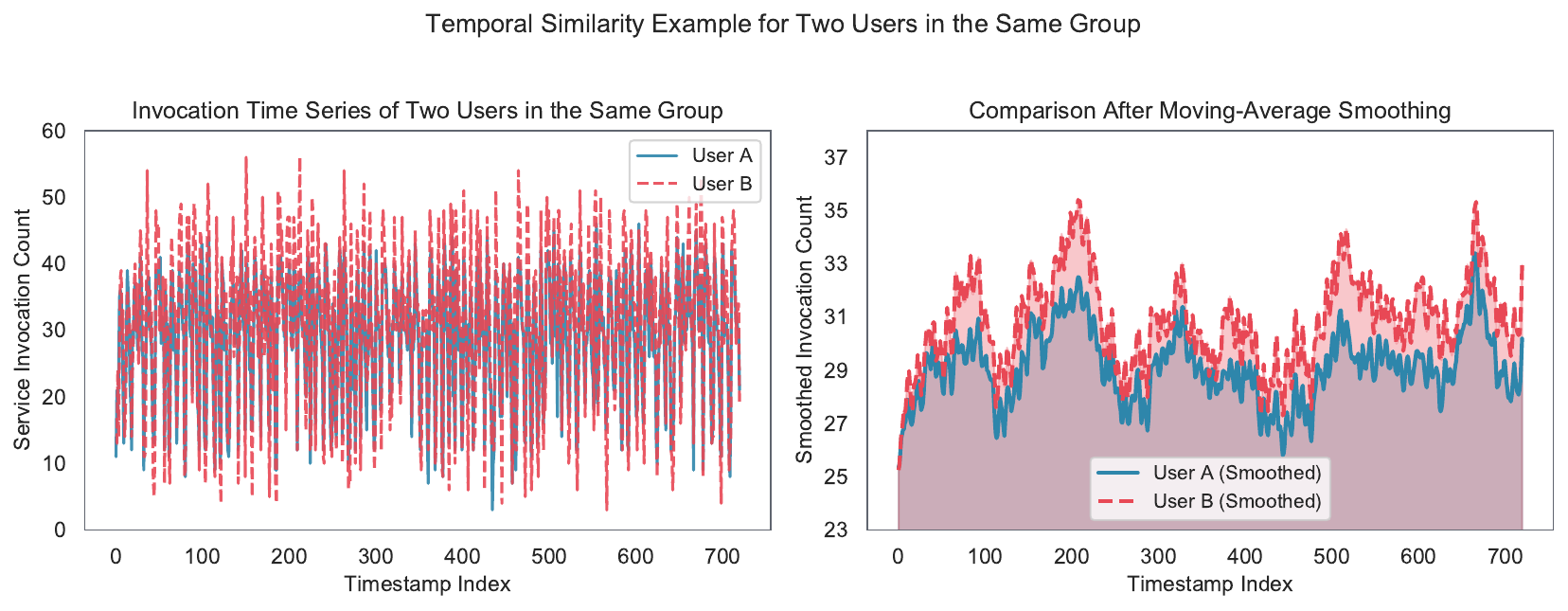}
\caption{Example of temporal similarity between two users from the same request-pattern group.}
\label{fig:figure7}
\end{figure}

\textbf{Temporal similarity within request groups.} The request synthesis module is intended to generate users with similar global trends inside the same group while preserving local variation at the individual level. Figure \ref{fig:figure7} illustrates one such pair of users.
Their raw invocation counts fluctuate differently at some timestamps, yet the smoothed curves remain highly consistent in terms of major rises and falls. This suggests that the grouped Gaussian-process construction captures shared behavioral rhythm without collapsing all users into identical traces.

\subsection{QoS-Related Statistics}
\begin{figure}[H]
\centering
\includegraphics[width=\columnwidth]{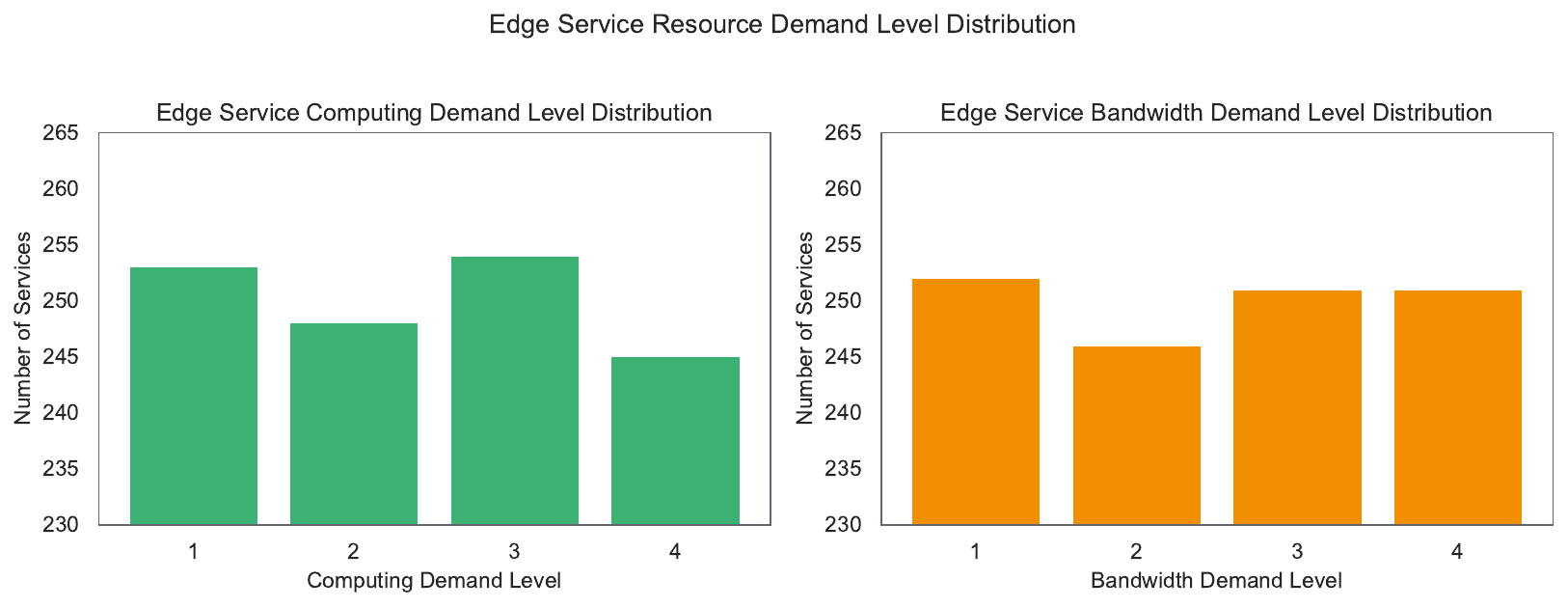}
\caption{Demand-level distributions of synthetic services.}
\label{fig:figure8}
\end{figure}

\textbf{Service diversity.} Figure \ref{fig:figure8} shows the demand-level distributions of the 1,000 synthetic services over computation, storage, and bandwidth.
The three histograms are close to uniform, with roughly 250 services per level. This balance prevents the dataset from being dominated by either lightweight or heavyweight service types and provides a wide range of operating conditions for QoS prediction.

\begin{figure}[H]
\centering
\includegraphics[width=\columnwidth]{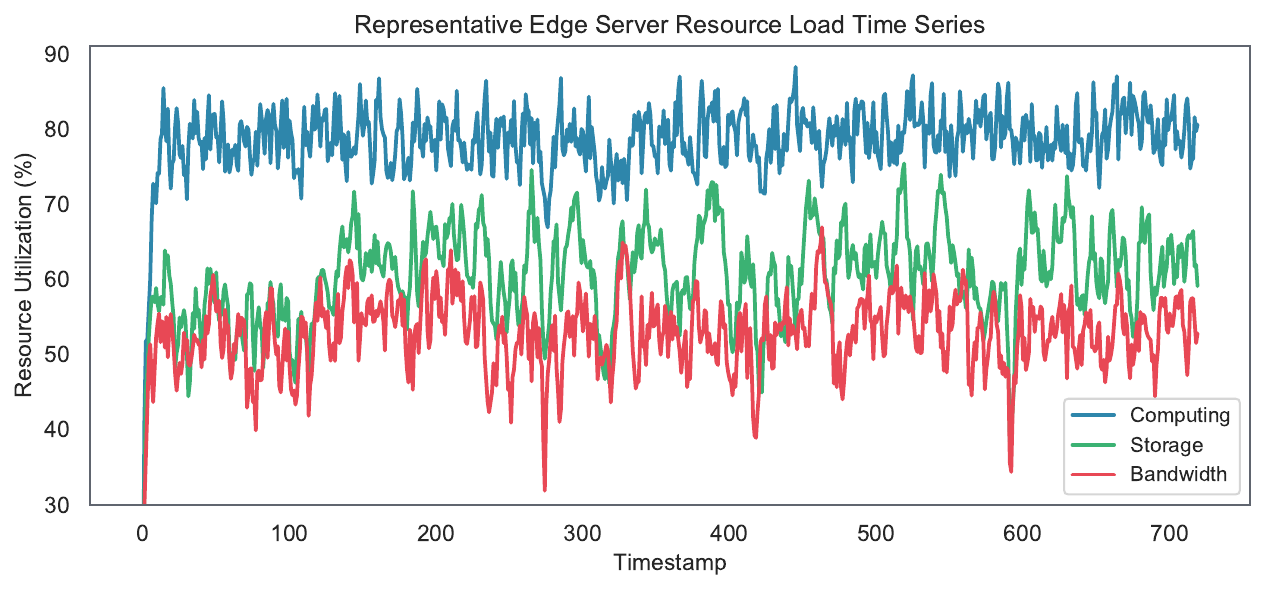}
\caption{Example of temporal load evolution for a representative edge server.}
\label{fig:figure9}
\end{figure}

\textbf{Temporal server load dynamics.} Figure \ref{fig:figure9} presents the time series of a representative server.
The mean utilization of computation, storage, and bandwidth is approximately 78.6\%, 60.5\%, and 52.8\%, respectively. More importantly, the curves evolve smoothly instead of repeatedly jumping between idle and saturated states. This indicates that the recursive load update with nonlinear release successfully preserves temporal dependence, which is crucial for forecasting tasks that rely on recent server history.

\begin{figure}[H]
\centering
\includegraphics[width=\columnwidth]{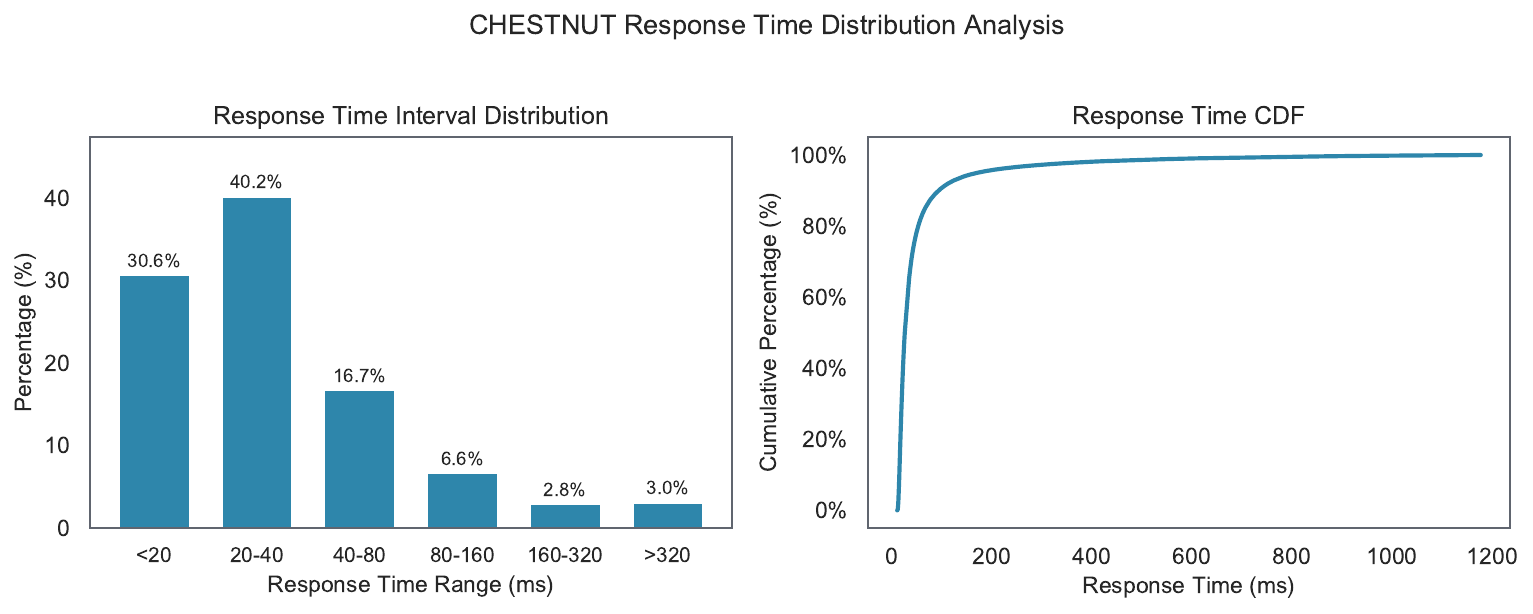}
\caption{Probability density and cumulative distribution of response time in CHESTNUT.}
\label{fig:figure10}
\end{figure}

\textbf{Response time distribution.} Figure \ref{fig:figure10} reports both the empirical density and the cumulative distribution of response time.
About 31.2\% of invocations are below 20 ms, and another 40.0\% lie between 20 ms and 40 ms, so 71.2\% of the records fall into a low-latency regime. The mean response time is about 60.9 ms, whereas the median is 25.8 ms, which indicates a clear right-skewed distribution with a moderate long tail. This behavior is desirable because it covers both common low-latency cases and rarer congested situations.

\begin{figure}[H]
\centering
\includegraphics[width=\columnwidth]{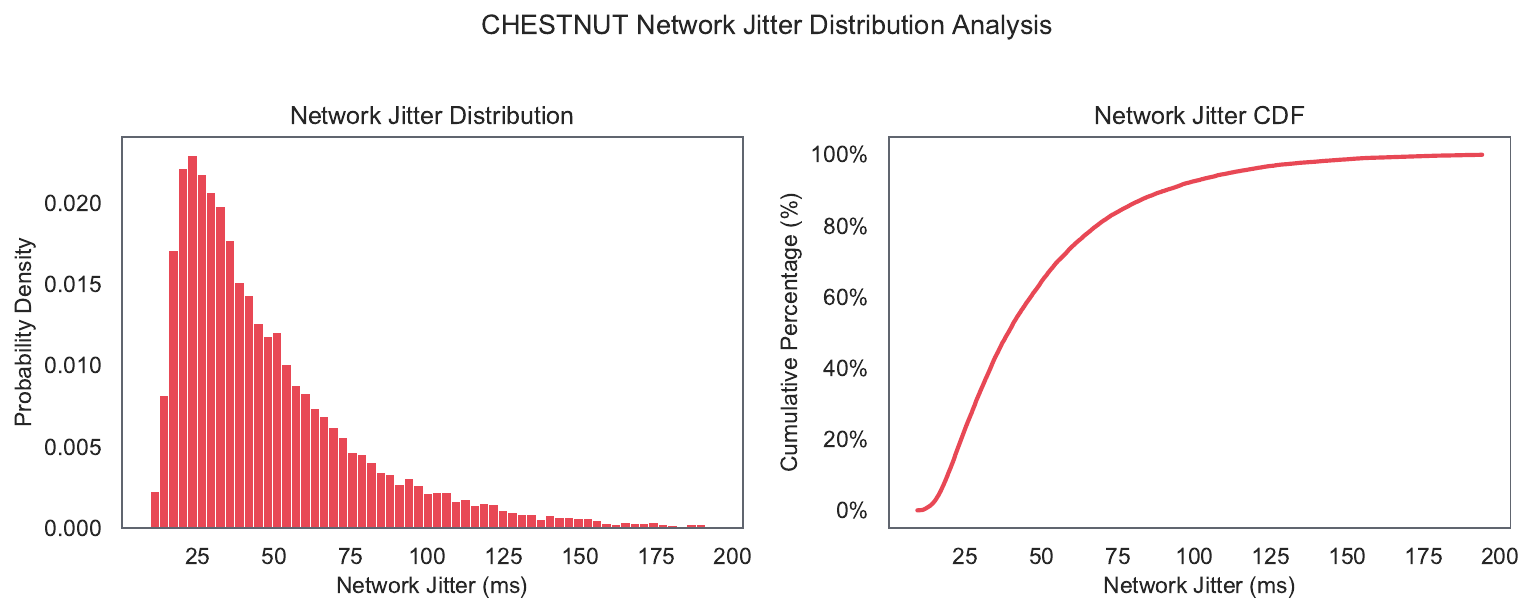}
\caption{Probability density and cumulative distribution of network jitter in CHESTNUT.}
\label{fig:figure11}
\end{figure}

\textbf{Network jitter distribution.} Figure \ref{fig:figure11} shows the distribution of network jitter.
Around 51.7\% of invocation records are below 40 ms, and 34.6\% lie between 40 ms and 80 ms. The mean and median are 48.7 ms and 39.0 ms, respectively. Compared with response time, jitter values are more concentrated in the middle range, which is consistent with the fact that most services experience relatively stable transmission while a smaller fraction suffers from stronger mobility- or bandwidth-driven fluctuation.

\bibliographystyle{IEEEtranBST/IEEEtran}
\bibliography{IEEEtranBST/IEEEexample}

@article{li2021profit,
  title={Profit-aware edge server placement},
  author={Li, Yuanzhe and Zhou, Ao and Ma, Xiao and Wang, Shangguang},
  journal={IEEE Internet of Things Journal},
  volume={9},
  number={1},
  pages={55--67},
  year={2021},
  publisher={IEEE}
}

@article{guo2020user,
  title={User allocation-aware edge cloud placement in mobile edge computing},
  author={Guo, Yan and Wang, Shangguang and Zhou, Ao and Xu, Jinliang and Yuan, Jie and Hsu, Ching-Hsien},
  journal={Software: Practice and Experience},
  volume={50},
  number={5},
  pages={489--502},
  year={2020},
  publisher={Wiley Online Library}
}

@article{wang2019delay,
  title={Delay-aware microservice coordination in mobile edge computing: A reinforcement learning approach},
  author={Wang, Shangguang and Guo, Yan and Zhang, Ning and Yang, Peng and Zhou, Ao and Shen, Xuemin},
  journal={IEEE Transactions on Mobile Computing},
  volume={20},
  number={3},
  pages={939--951},
  year={2019},
  publisher={IEEE}
}

@article{kritikos2009requirements,
  title={Requirements for QoS-based web service description and discovery},
  author={Kritikos, Kyriakos and Plexousakis, Dimitris},
  journal={IEEE Transactions on Services Computing},
  volume={2},
  number={4},
  pages={320--337},
  year={2009},
  publisher={IEEE}
}

@article{karakus2017quality,
  title={Quality of service (QoS) in software defined networking (SDN): A survey},
  author={Karakus, Murat and Durresi, Arjan},
  journal={Journal of Network and Computer Applications},
  volume={80},
  pages={200--218},
  year={2017},
  publisher={Elsevier}
}

@inproceedings{xue2020edge,
  title={Edge computing for internet of things: A survey},
  author={Xue, Huihui and Huang, Bi and Qin, Mingming and Zhou, Hua and Yang, Hongji},
  booktitle={2020 International Conferences on Internet of Things (iThings) and IEEE Green Computing and Communications (GreenCom) and IEEE Cyber, Physical and Social Computing (CPSCom) and IEEE Smart Data (SmartData) and IEEE Congress on Cybermatics (Cybermatics)},
  pages={755--760},
  year={2020},
  organization={IEEE}
}

@article{karimi2017qos,
  title={QoS-aware service composition in cloud computing using data mining techniques and genetic algorithm},
  author={Karimi, Mohammad Bagher and Isazadeh, Ayaz and Rahmani, Amir Masoud},
  journal={The Journal of Supercomputing},
  volume={73},
  pages={1387--1415},
  year={2017},
  publisher={Springer}
}

@article{syu2017time,
  title={Time series forecasting for dynamic quality of web services: An empirical study},
  author={Syu, Yang and Kuo, Jong-Yih and Fanjiang, Yong-Yi},
  journal={Journal of Systems and Software},
  volume={134},
  pages={279--303},
  year={2017},
  publisher={Elsevier}
}

@Article{electronics13163113,
AUTHOR = {Cristobo, Leire and Ibarrola, Eva and Casado-O’Mara, Itziar and Zabala, Luis},
TITLE = {Global Quality of Service (QoX) Management for Wireless Networks},
JOURNAL = {Electronics},
VOLUME = {13},
YEAR = {2024},
NUMBER = {16},
ARTICLE-NUMBER = {3113},
URL = {https://www.mdpi.com/2079-9292/13/16/3113},
ISSN = {2079-9292},
DOI = {10.3390/electronics13163113}
}

@article{Menasc2002QoSII,
  title={QoS Issues in Web Services},
  author={Daniel A. Menasc{\'e}},
  journal={IEEE Internet Comput.},
  year={2002},
  volume={6},
  pages={72-75},
  url={https://api.semanticscholar.org/CorpusID:42937779}
}

@inproceedings{Shade2012QualityOS,
  title={Quality of Service (Qos) Issues in Web Services},
  author={Kuyoro Shade and Awodele O. Akinde and Osunlana Ronke and O Okolie Samuel},
  year={2012},
  url={https://api.semanticscholar.org/CorpusID:167434500}
}

@article{Wu2013PredictingQO,
  title={Predicting Quality of Service for Selection by Neighborhood-Based Collaborative Filtering},
  author={Jian Wu and Liang Chen and Yi Feng and Zibin Zheng and Mengchu Zhou and Zhaohui Wu},
  journal={IEEE Transactions on Systems, Man, and Cybernetics: Systems},
  year={2013},
  volume={43},
  pages={428-439},
  url={https://api.semanticscholar.org/CorpusID:15261484}
}

@article{Li2017TemporalIC,
  title={Temporal Influences-Aware Collaborative Filtering for QoS-Based Service Recommendation},
  author={Jinglin Li and Jie Wang and Qibo Sun and Ao Zhou},
  journal={2017 IEEE International Conference on Services Computing (SCC)},
  year={2017},
  pages={471-474},
  url={https://api.semanticscholar.org/CorpusID:25892490}
}

@article{Zheng2020WebSQ,
  title={Web Service QoS Prediction via Collaborative Filtering: A Survey},
  author={Zibin Zheng and Liao Xiaoli and Mingdong Tang and Fenfang Xie and Michael R. Lyu},
  journal={IEEE Transactions on Services Computing},
  year={2020},
  volume={15},
  pages={2455-2472},
  url={https://api.semanticscholar.org/CorpusID:219501848}
}

@article{Wu2016TimeAwareAS,
  title={Time-Aware and Sparsity-Tolerant QoS Prediction Based on Collaborative Filtering},
  author={Chen Wu and Weiwei Qiu and Xinyu Wang and Zibin Zheng and Xiaohu Yang},
  journal={2016 IEEE International Conference on Web Services (ICWS)},
  year={2016},
  pages={637-640},
  url={https://api.semanticscholar.org/CorpusID:837796}
}

@article{Tang2015CloudSQ,
  title={Cloud service QoS prediction via exploiting collaborative filtering and location‐based data smoothing},
  author={Mingdong Tang and Tingting Zhang and Jianxun Liu and Jinjun Chen},
  journal={Concurrency and Computation: Practice and Experience},
  year={2015},
  volume={27},
  pages={5826 - 5839},
  url={https://api.semanticscholar.org/CorpusID:24993426}
}

@article{Su2016WebSQ,
  title={Web service QoS prediction by neighbor information combined non-negative matrix factorization},
  author={Kai Su and Liangli Ma and Bin Xiao and Huaiqiang Zhang},
  journal={J. Intell. Fuzzy Syst.},
  year={2016},
  volume={30},
  pages={3593-3604},
  url={https://api.semanticscholar.org/CorpusID:29886978}
}

@article{Zhang2011CollaborativeFB,
  title={Collaborative Filtering Based Service Ranking Using Invocation Histories},
  author={Qiong Zhang and Chen Ding and Chi-Hung Chi},
  journal={2011 IEEE International Conference on Web Services},
  year={2011},
  pages={195-202},
  url={https://api.semanticscholar.org/CorpusID:15071119}
}

@article{Wu2017CollaborativeQP,
  title={Collaborative QoS prediction with context-sensitive matrix factorization},
  author={Hao Wu and Kun Yue and Bo Li and Binbin Zhang and Ching-Hsien Hsu},
  journal={Future Gener. Comput. Syst.},
  year={2017},
  volume={82},
  pages={669-678},
  url={https://api.semanticscholar.org/CorpusID:3637653}
}

@article{Zhou2023SpatialCT,
  title={Spatial Context-Aware Time-Series Forecasting for QoS Prediction},
  author={Jie Zhou and Ding Ding and Ziteng Wu and Yuting Xiu},
  journal={IEEE Transactions on Network and Service Management},
  year={2023},
  volume={20},
  pages={918-931},
  url={https://api.semanticscholar.org/CorpusID:257300058}
}

@article{Zhang2023PredictingQO,
  title={Predicting Quality of Services Based on a Two-Stream Deep Learning Model With User and Service Graphs},
  author={Peiyun Zhang and Wenjun Huang and Yutong Chen and Mengchu Zhou},
  journal={IEEE Transactions on Services Computing},
  year={2023},
  volume={16},
  pages={4060-4072},
  url={https://api.semanticscholar.org/CorpusID:261149398}
}

@article{Jin2019NeighborhoodawareWS,
  title={Neighborhood-aware web service quality prediction using deep learning},
  author={Ying Jin and Kaibin Wang and Yiwen Zhang and Yuanting Yan},
  journal={EURASIP Journal on Wireless Communications and Networking},
  year={2019},
  volume={2019},
  pages={1-10},
  url={https://api.semanticscholar.org/CorpusID:201838513}
}

@inproceedings{Awanyo2023DeepNN,
  title={Deep Neural Network-Based Approach for IoT Service QoS Prediction},
  author={Christson Awanyo and Nawal Guermouche},
  booktitle={WISE},
  year={2023},
  url={https://api.semanticscholar.org/CorpusID:264441803}
}

@article{graves2012long,
  title={Long short-term memory},
  author={Graves, Alex and Graves, Alex},
  journal={Supervised sequence labelling with recurrent neural networks},
  pages={37--45},
  year={2012},
  publisher={Springer}
}

@inproceedings{he2016deep,
  title={Deep residual learning for image recognition},
  author={He, Kaiming and Zhang, Xiangyu and Ren, Shaoqing and Sun, Jian},
  booktitle={Proceedings of the IEEE conference on computer vision and pattern recognition},
  pages={770--778},
  year={2016}
}

@article{zou2022ncrl,
  title={NCRL: Neighborhood-based collaborative residual learning for adaptive QoS prediction},
  author={Zou, Guobing and Wu, Shaogang and Hu, Shengxiang and Cao, Chenhong and Gan, Yanglan and Zhang, Bofeng and Chen, Yixin},
  journal={IEEE Transactions on Services Computing},
  volume={16},
  number={3},
  pages={2030--2043},
  year={2022},
  publisher={IEEE}
}

@article{Koursioumpas2024DISTINQTAD,
  title={DISTINQT: A Distributed Privacy Aware Learning Framework for QoS Prediction for Future Mobile and Wireless Networks},
  author={Nikolaos Koursioumpas and Lina Magoula and Ioannis Stavrakakis and Nancy Alonistioti and Miguel Angel Gutierrez-Estevez and Ramin Khalili},
  journal={ArXiv},
  year={2024},
  volume={abs/2401.10158},
  url={https://api.semanticscholar.org/CorpusID:267035178}
}

@article{Zou2023FHCDQPFH,
  title={FHC-DQP: Federated Hierarchical Clustering for Distributed QoS Prediction},
  author={Guobing Zou and Shiyi Lin and Shengxiang Hu and Shengyu Duan and Yanglan Gan and Bofeng Zhang and Yixin Chen},
  journal={IEEE Transactions on Services Computing},
  year={2023},
  volume={16},
  pages={4073-4086},
  url={https://api.semanticscholar.org/CorpusID:261310990}
}

@article{Zou2024TRCFTR,
  title={TRCF: Temporal Reinforced Collaborative Filtering for Time-Aware QoS Prediction},
  author={Guobing Zou and Yutao Huang and Shengxiang Hu and Yanglan Gan and Bofeng Zhang and Yixin Chen},
  journal={IEEE Transactions on Services Computing},
  year={2024},
  volume={17},
  pages={1847-1860},
  url={https://api.semanticscholar.org/CorpusID:265124546}
}

@article{zou2024frln,
  title={FRLN: Federated Residual Ladder Network for Data-Protected QoS Prediction},
  author={Zou, Guobing and Yu, Wenzhuo and Hu, Shengxiang and Gan, Yanglan and Zhang, Bofeng and Chen, Yixin},
  journal={IEEE Transactions on Services Computing},
  year={2024},
  publisher={IEEE}
}

@article{zheng2012investigating,
  title={Investigating QoS of real-world web services},
  author={Zheng, Zibin and Zhang, Yilei and Lyu, Michael R},
  journal={IEEE transactions on services computing},
  volume={7},
  number={1},
  pages={32--39},
  year={2012},
  publisher={IEEE}
}

@article{Soldani20145GNE,
  title={5G networks: End-to-end architecture and infrastructure [Guest Editorial]},
  author={David Soldani and Kostas Pentikousis and Rahim Tafazolli and Daniele Franceschini},
  journal={IEEE Commun. Mag.},
  year={2014},
  volume={52},
  pages={62-64},
  url={https://api.semanticscholar.org/CorpusID:33738654}
}

@article{Chen2008AnalysisOW,
  title={Analysis of web response time in asymmetrical wireless network},
  author={Xiaohui Chen and Weidong Wang and Jincheng Nie},
  journal={2008 11th IEEE Singapore International Conference on Communication Systems},
  year={2008},
  pages={1427-1430},
  url={https://api.semanticscholar.org/CorpusID:43052448}
}

@article{Ruan2017RoundTripDM,
  title={Round-Trip Delay Modeling for Smart Body Area Networks},
  author={Lihua Ruan and Maluge Pubuduni Imali Dias and Ye Feng and Elaine Wong},
  journal={IEEE Communications Letters},
  year={2017},
  volume={21},
  pages={2528-2531},
  url={https://api.semanticscholar.org/CorpusID:4923183}
}

@article{Ulvan2009DelayPO,
  title={Delay Performance of Session Establishment Signaling in IP Multimedia Subsystem},
  author={Melvi Ulvan and Robert Bestak},
  journal={2009 16th International Conference on Systems, Signals and Image Processing},
  year={2009},
  pages={1-5},
  url={https://api.semanticscholar.org/CorpusID:18626127}
}

@article{jin2024swift,
  author  = {Huiying Jin and
             Pengcheng Zhang and
             Hai Dong and
             Athman Bouguettaya and
             Albert Y. Zomaya},
  title   = {Swift and Accurate Mobility-Aware {QoS} Forecasting for Mobile Edge
             Environments},
  journal = {IEEE Transactions on Services Computing},
  volume  = {17},
  number  = {6},
  pages   = {4340--4353},
  year    = {2024}
}

@article{chen2023mobility,
  title     = {Mobility-aware edge server placement for mobile edge computing},
  author    = {Chen, Yuanyi and Wang, Dezhi and Wu, Nailong and Xiang, Zhengzhe},
  journal   = {Computer Communications},
  volume    = {208},
  pages     = {136--146},
  year      = {2023},
  publisher = {Elsevier}
}

@book{rasmussen2006gaussian,
  author    = {Carl Edward Rasmussen and
               Christopher K. I. Williams},
  title     = {Gaussian Processes for Machine Learning},
  publisher = {{MIT} Press},
  year      = {2006}
}

@book{kleinrock1975queueing,
  author    = {Leonard Kleinrock},
  title     = {Queueing Systems, Volume 1: Theory},
  publisher = {Wiley},
  year      = {1975}
}

@book{kurose2017computer,
  author    = {James F. Kurose and
               Keith W. Ross},
  title     = {Computer Networking: A Top-Down Approach},
  edition   = {7},
  publisher = {Pearson},
  year      = {2017}
}

@techreport{rfc3393,
  author      = {C. Demichelis and
                 P. Chimento},
  title       = {{IP} Packet Delay Variation Metric for {IP} Performance Metrics
                 ({IPPM})},
  institution = {{RFC} Editor},
  type        = {{RFC}},
  number      = {3393},
  year        = {2002},
  url         = {https://www.rfc-editor.org/rfc/rfc3393}
}

%
\vspace{-1.5cm}
\begin{IEEEbiography}[{\includegraphics[width=1in,height=1.25in,clip,keepaspectratio]{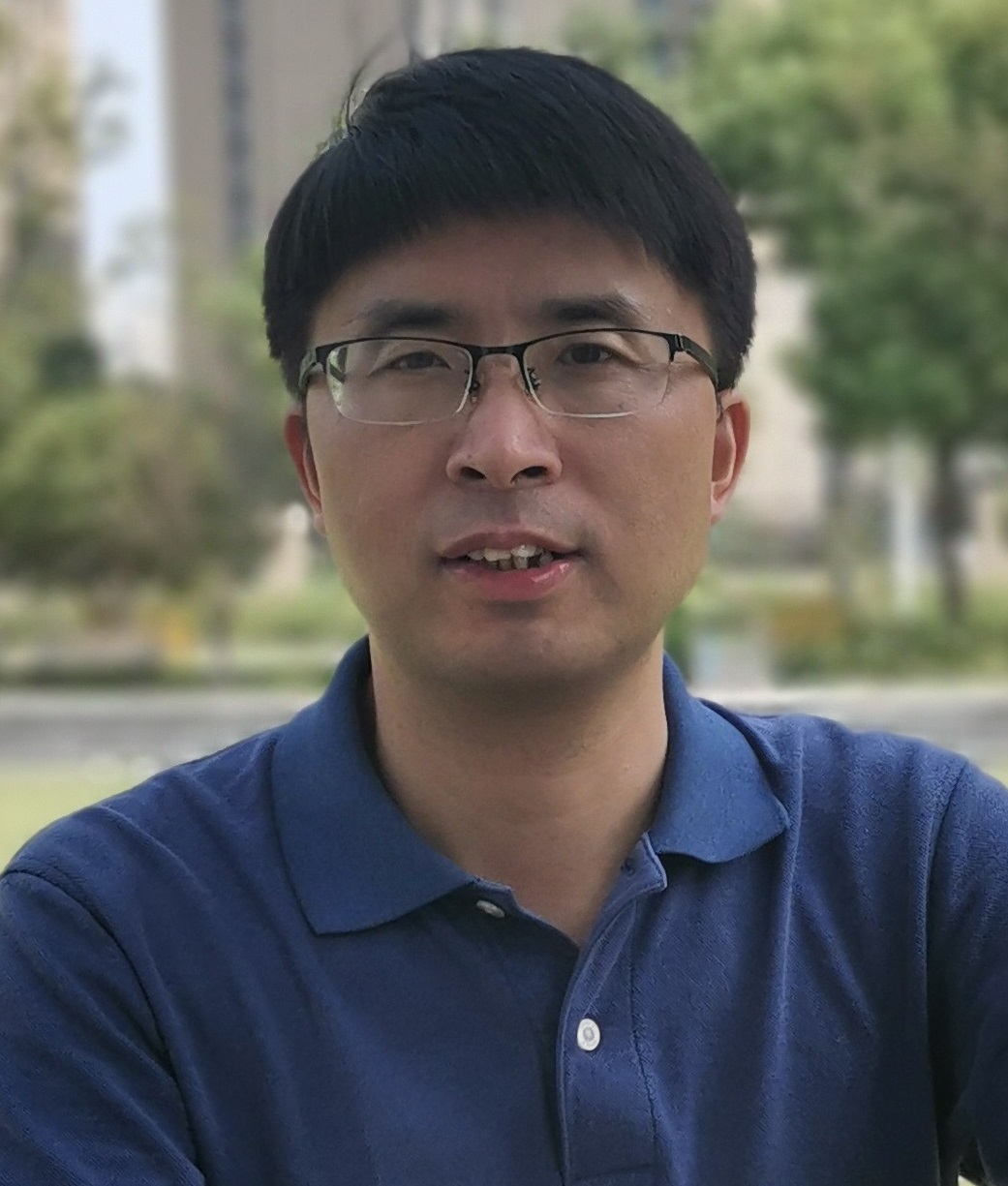}}]{Guobing Zou}
	 is a full professor and dean of the Department of Computer Science and Technology, Shanghai University, China. He received his PhD degree in Computer Science from Tongji University, Shanghai, China, 2012. He has worked as a visiting scholar in the Department of Computer Science and Engineering at Washington University in St. Louis from 2009 to 2011, USA. His current research interests mainly focus on services computing, edge computing, data mining and intelligent algorithms, recommender systems. He has published more than 100 papers on premier international journals and conferences, including IEEE Transactions on Services Computing, IEEE Transactions on Network and Service Management, IEEE ICWS, ICSOC, IEEE SCC, AAAI, IJWSR, IJWGS, Information Sciences, Expert Systems with Applications, Knowledge- Based Systems, Applied Intelligence, etc. He served as organization and publicity chair of the International Conference on Service Science, vice chair of IEEE International Conference on Big Data (IEEE BigData 2021), chair of ``Service Computing Top Conference Top Journal Forum'' of China Digital Service Conference (2021-2023), and guest editor of International Journal of Services Technology and Management.
\end{IEEEbiography}
\vspace{-1.3cm}
\begin{IEEEbiography}
	[{\includegraphics[width=1in,height=1.25in,clip,keepaspectratio]{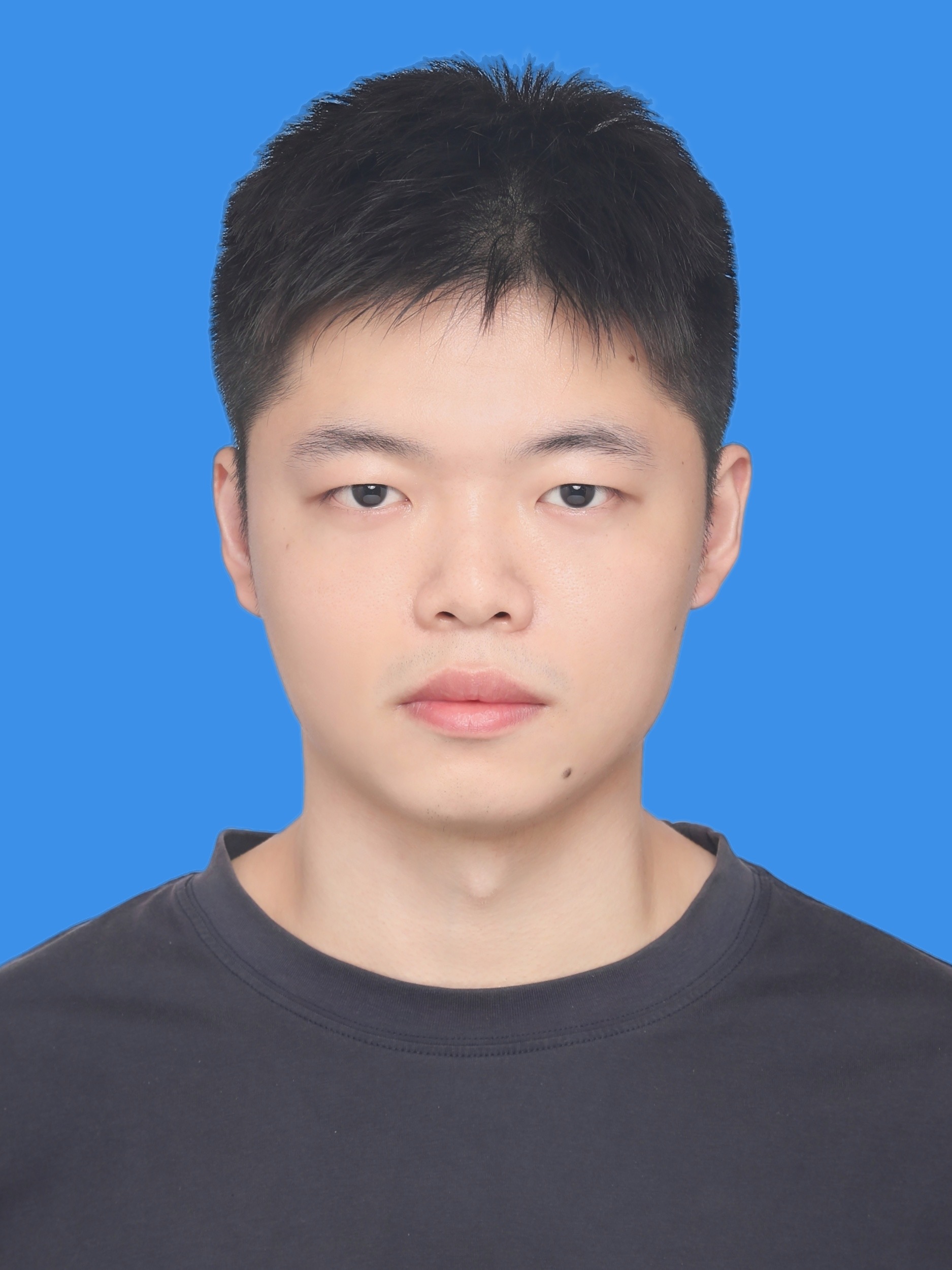}}]{Fei Zhao}
	received the bachelor’s degree in software engineering from Fujian Normal University, China, 2022. He is currently working toward the master’s degree with the School of Computer Engineering and Science, Shanghai University, China. His research interests include QoS prediction in mobile edge environment, deep learning, graph neural network. His current work involves the application of spatio-temporal graph neural networks to enhance the accuracy of QoS predictions in dynamic mobile edge scenarios.
\vspace{-1.3cm}
\end{IEEEbiography}
\begin{IEEEbiography}
	[{\includegraphics[width=1in,height=1.25in,clip,keepaspectratio]{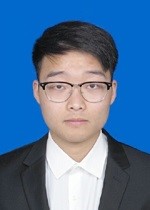}}]{Shengxiang Hu}
	received the bachelor degree in 2018 and the master degree in 2021 both in computer science and Technology from Shanghai University, respectively. He is currently working toward the PhD degree in the School of Computer Engineering and Science, Shanghai University, China. His research interests include QoS prediction, graph neural network and natural language processing. He has published more than five papers on IEEE Transactions on Services Computing, IEEE Knowledge-Based Systems, International Conference on Web Services (IEEE ICWS), International Conference on Service-Oriented Computing (ICSOC), International Conference on Parallel Problem Solving from Nature (PPSN).
\end{IEEEbiography}
\end{document}